\documentclass[lettersize,journal]{IEEEtran}
\usepackage{amsmath,amsfonts}
\usepackage{algorithmic}
\usepackage{algorithm}
\usepackage{array}
\usepackage[caption=false,font=normalsize,labelfont=sf,textfont=sf]{subfig}
\usepackage{textcomp}
\usepackage{stfloats}
\usepackage{url}
\usepackage{verbatim}
\usepackage{graphicx}
\usepackage{pifont}
\usepackage{cite}
\usepackage{float}
\usepackage{booktabs}
\usepackage{multirow}
\usepackage[table,xcdraw]{xcolor}
\begin{document}

\title{DistMedVL: Distributional Vision-Language Alignment for Uncertainty-Aware Medical Image Segmentation}

\author{Jiaxuan Li, Qing Xu, Xiangjian He,~\IEEEmembership{Senior Member,~IEEE,} Yue Li, Daokun Zhang, Fiseha B. Tesema, Rong Qu,~\IEEEmembership{Fellow,~IEEE}

\thanks{\quad This work is partially supported by the Yongjiang Technology Innovation Project (2022A-097-G), Zhejiang Department of Transportation General Research and Development Project (2024039), and National Natural Science Foundation of China grant (UNNC: B0166). \textit{(Equal contribution: Jiaxuan Li and Qing Xu, Corresponding authors: Xiangjian He)}}
\thanks{\quad J. Li and Q. Xu are with School of Computer Science, University of Nottingham Ningbo China, Ningbo, Zhejiang, China, and with School of Computer Science, University of Nottingham, UK (e-mail: jiaxuan.li@nottingham.edu.cn).}
\thanks{\quad X. He, Y. Li, F. B. Tesem, D. Zhang are with School of Computer Science, University of Nottingham Ningbo China, Ningbo, Zhejiang, China (e-mail: sean.he@nottingham.edu.cn).}
\thanks{\quad R. Qu is with School of Computer Science, University of Nottingham, UK (e-mail: rong.qu@nottingham.ac.uk).}}

\markboth{Journal of \LaTeX\ Class Files,~Vol.~14, No.~8, August~2021}%
{Shell \MakeLowercase{\textit{et al.}}: A Sample Article Using IEEEtran.cls for IEEE Journals}

\IEEEpubid{}

\maketitle

\begin{abstract}
Cross-modal alignment of visual and textual representations is fundamental to multimodal medical image understanding, yet remains hindered by uncertainty in both modalities under real-world clinical conditions. Existing vision-language segmentation methods rely on deterministic cross-modal matching, which overlooks aleatoric uncertainty from ambiguous boundaries and epistemic uncertainty from limited training data, leading to fragile performance under domain shift. To address this issue, we propose DistMedVL, a probabilistic vision-language framework that introduces a lightweight Probabilistic Cross-Modal Adapter (PCM-Adapter) upon frozen encoders to explicitly model representational uncertainty. Specifically, the PCM-Adapter comprises two sequential modules for progressive probabilistic alignment. We first devise a Mahalanobis Alignment Module (MAM) that models textual tokens as Gaussian distributions and computes patch-text compatibility via Mahalanobis distance, yielding variance-conditioned matching that downweights unreliable feature dimensions. Moreover, we devise a Distribution Flow Module (DFM) that estimates modality-wise confidence parameters and performs vision-guided refinement of textual distributions, accommodating distributional variation across imaging modalities. Extensive experiments across eight medical segmentation benchmarks demonstrate that DistMedVL outperforms state-of-the-art methods with only 6.3M trainable parameters, exhibiting superior data efficiency, perturbation robustness and cross-dataset generalization.
\end{abstract}

\section{Introduction}
\IEEEPARstart{M}{edical} image segmentation is fundamental to clinical applications such as computer-aided diagnosis, treatment planning and surgical navigation \cite{litjens2017survey, anwar2018medical, shurrab2022self}. In clinical practice, medical images are inherently accompanied by rich textual information, including radiology reports, diagnostic descriptions and clinical notes, which provide complementary semantic context about anatomical structures and pathological conditions. This naturally multimodal clinical workflow motivates cross-modal learning paradigms that jointly leverage visual and textual modalities 
to improve the accuracy and generalizability of medical image segmentation \cite{wang2022medclip}. Such paradigms are particularly beneficial for medical segmentation, where textual descriptions of lesion characteristics, locations, and pathological findings can provide essential semantic guidance for delineating structures that are visually ambiguous across different imaging modalities.
 
Early medical image segmentation methods primarily focused on unimodal visual learning. CNN-based architectures \cite{ronneberger2015u, wu2022fat} established foundational feature extraction capabilities through hierarchical local pattern learning, while Transformer-based approaches \cite{cao2022swin, azad2022transdeeplab} introduced global self-attention mechanisms for capturing long-range dependencies. Hybrid architectures \cite{chen2021transunet, he2023h2former} further combined the strengths of both paradigms, achieving enhanced pixel-level prediction across various imaging modalities. Despite their success, these unimodal methods are inherently confined to visual feature representations, leaving the potential of language-based guidance for more flexible segmentation unexplored. To exploit the complementary textual information available in clinical settings, vision-language models (VLMs), pre-trained on large-scale image-text pairs to learn aligned cross-modal representations \cite{radford2021learning}, have been increasingly adopted for medical image segmentation through strategies such as lightweight adapter-based transfer \cite{dhakal2024vlsm}, text-guided feature fusion \cite{li2023lvit, yang2022lavt}, and cross-modal attention mechanisms \cite{rao2022denseclip}.

\begin{figure*}[!t]
\centering
\includegraphics[width=\textwidth]{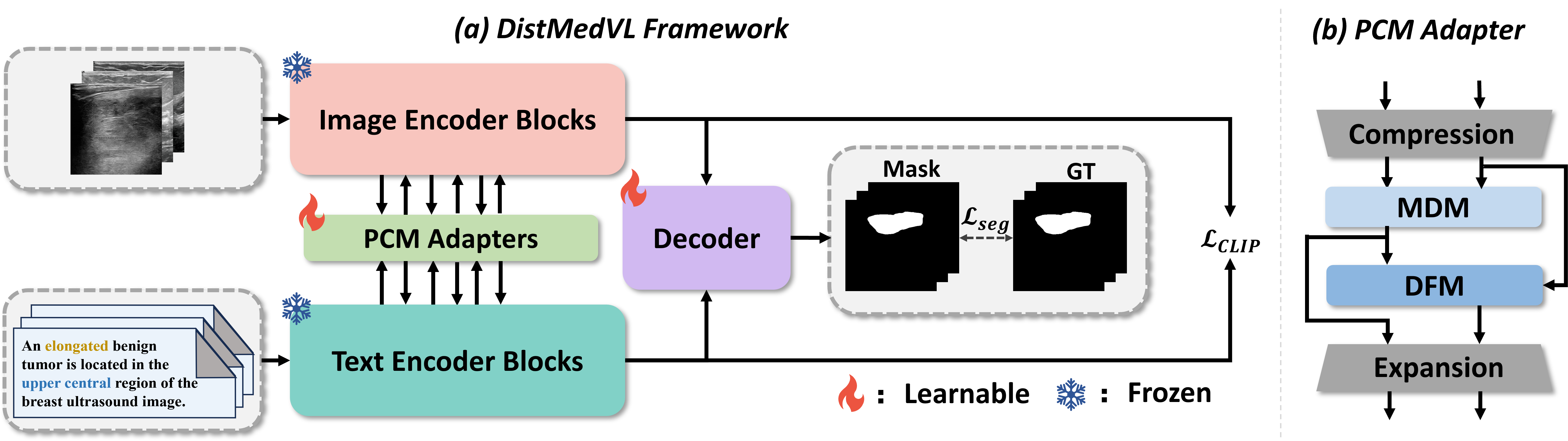}
\caption{(a) Overview of the DistMedVL architecture, where only the PCM adapter and decoder are trainable. (b) Detail of the PCM Adapter, which maps cross-modal features to probabilistic embeddings.}
\label{fig:architecture}
\end{figure*}

While these vision-language approaches achieve promising performance, their deterministic cross-modal alignment paradigm remains inherently fragile under domain shift and data-scarce scenarios. In real-world clinical settings, image quality is frequently degraded by device heterogeneity and acquisition noise \cite{wen2024denoising}, while paired textual inputs often contain semantic ambiguity or erroneous annotations \cite{huang2024enhancing}. Furthermore, the scarcity of large-scale annotated medical datasets \cite{TAJBAKHSH2020101693} amplifies model uncertainty in underrepresented regions and rare pathologies. Under such conditions, deterministic alignment treats visual and textual representations as fixed-point embeddings without assessing per-dimension feature reliability \cite{xu2024deterministic}, overlooking both the aleatoric uncertainty inherent in noisy inputs and the epistemic uncertainty arising from limited training data. These limitations motivate the need for probabilistic modeling in cross-modal alignment.
 
To address this issue, we propose DistMedVL, a probabilistic vision-language framework that introduces a lightweight Probabilistic Cross-Modal Adapter (PCM-Adapter) upon frozen vision-language encoders to explicitly model representational uncertainty. The PCM-Adapter comprises two sequential modules for progressive cross-modal probabilistic alignment. The Mahalanobis Alignment Module (MAM) models textual tokens as Gaussian distributions and computes patch-text compatibility via Mahalanobis distance, where bilateral variances estimated from both modalities modulate the matching scores to naturally downweight unreliable feature dimensions, enabling anisotropic alignment that accounts for both aleatoric and epistemic uncertainty. The Distribution Flow Module (DFM) estimates modality-wise confidence parameters and performs vision-guided refinement of textual distributions through a reliability-gated mechanism, enabling text representations to adaptively adjust based on visual evidence rather than serving as a fixed semantic prior. Together, the two modules form a progressive alignment pipeline that jointly addresses the uncertainty of both modalities. We comprehensively evaluate DistMedVL from both epistemic and aleatoric uncertainty perspectives, including data efficiency under limited supervision, cross-domain generalization under distribution shift, and perturbation robustness under degraded input quality. Extensive experiments on eight medical segmentation benchmarks demonstrate that DistMedVL consistently outperforms state-of-the-art methods with superior robustness under varying uncertainty conditions. The main contributions of this work are summarized as follows.
\begin{itemize}
    \item We propose DistMedVL, a probabilistic vision-language framework that introduces a lightweight PCM-Adapter into frozen vision-language encoders to jointly mitigate aleatoric and epistemic uncertainty in cross-modal medical image segmentation.
 
    \item We devise a Mahalanobis Alignment Module (MAM) that models textual tokens as Gaussian distributions and computes patch-text compatibility via Mahalanobis distance, replacing deterministic matching with variance-aware distribution-level alignment.
 
    \item We devise a Distribution Flow Module (DFM) that estimates modality-wise confidence parameters and employs a reliability gate to suppress high-variance visual features, enabling adaptive vision-guided refinement of textual distributions.
 
    \item Extensive experiments on eight medical segmentation benchmarks demonstrate that DistMedVL achieves state-of-the-art performance with only 6.3M trainable parameters under varying levels of uncertainty.
\end{itemize}

\section{Related Work}
\begin{figure*}[!t]
\centering
\includegraphics[width=\textwidth]{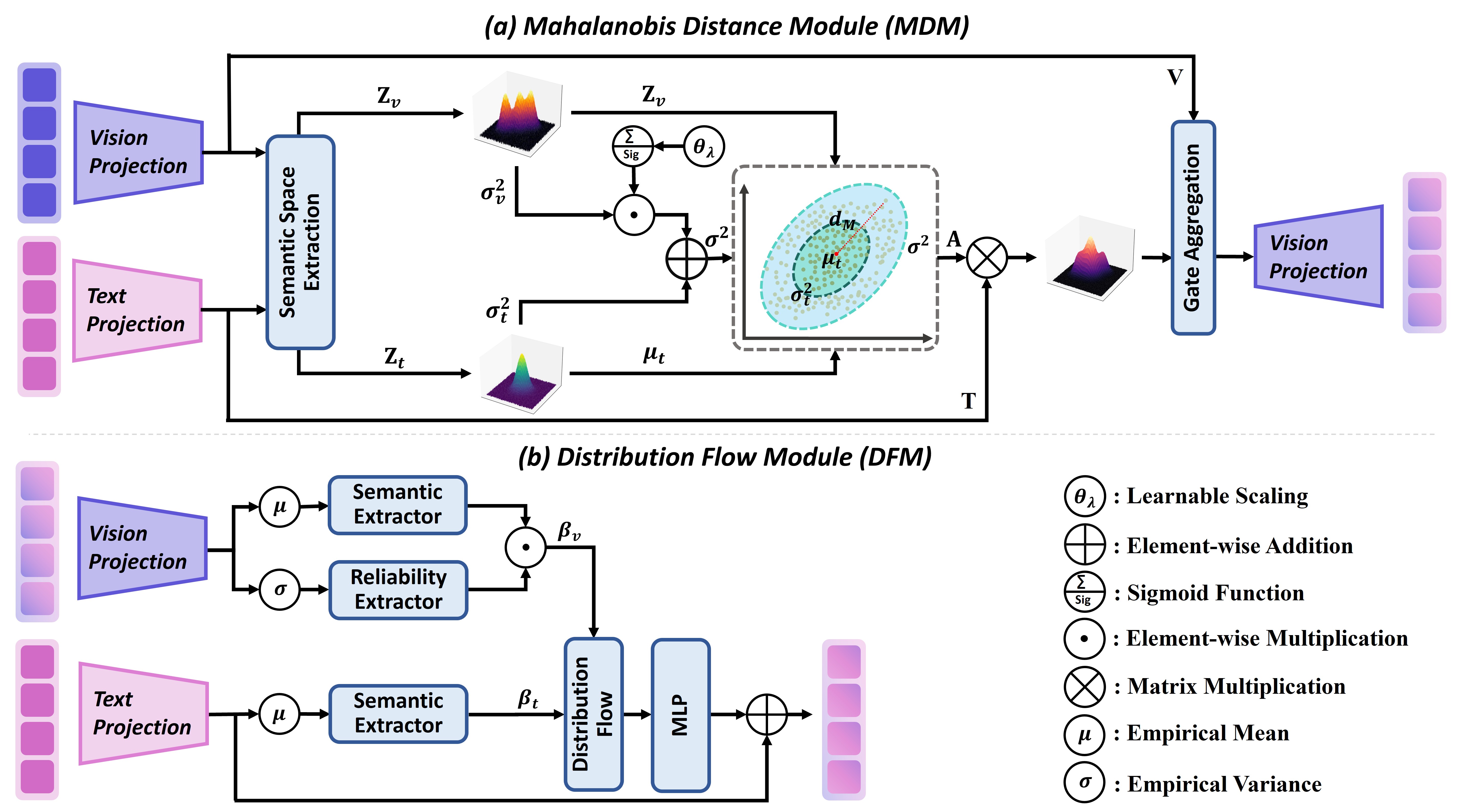}
\caption{Overall framework of the proposed method. (a) Mahalanobis Distance Module (MDM), which maps image features to text features. (b) Distribution Flow Module (DFM), which maps text features to global image semantics.}
\label{fig:framework}
\end{figure*}
\subsection{Unimodal Medical Image Segmentation}
Medical image segmentation has advanced significantly through deep learning architectures \cite{azad2024medical}. CNN-based methods centered on the U-Net framework \cite{ronneberger2015u} established foundational encoder-decoder designs, with subsequent works such as FatNet \cite{wu2022fat} and DCSAU-Net \cite{xu2023dcsau} enhancing multi-scale feature extraction and target awareness. To address the limited receptive field of convolutions, Transformer-based approaches such as Swin-UNet \cite{cao2022swin} introduced self-attention mechanisms for global context modeling, while hybrid CNN-Transformer architectures including TransUNet \cite{chen2021transunet}, H2Former \cite{he2023h2former}  and CFFormer \cite{li2026cfformer} combined local and global feature extraction for enhanced pixel-level prediction. Despite their strong performance across diverse imaging modalities, these methods produce deterministic representations that lack the capacity to model uncertainty, tending to yield over-confident predictions in data-scarce scenarios and out-of-distribution regions.
 
\subsection{Vision-Language Models for Medical Image Segmentation}
Vision-language models (VLMs) such as CLIP \cite{radford2021learning} learn aligned cross-modal representations from large-scale image-text pairs, demonstrating strong transferability across downstream tasks. This paradigm has been extended to the medical domain through models such as PubMedCLIP \cite{eslami2023pubmedclip} and UniMedCLIP \cite{khattak2024unimed}, which leverage domain-specific image-text pairs for medical representation learning. For medical image segmentation, a prevalent strategy introduces lightweight adapters upon frozen VLM encoders to achieve cross-modal segmentation at minimal computational cost. VLSM-Adapter \cite{dhakal2024vlsm} inserts adapter modules into both modality encoders but operates each modality independently without explicit cross-modal interaction. CausalCLIPSeg \cite{chen2024causalclipseg} employs causal intervention on visual features to suppress confounding factors, yet operates solely on visual representations and requires end-to-end fine-tuning with substantial computational overhead. MedCLIPSeg \cite{koleilat2026medclipseg} introduces probabilistic cross-modal alignment, yet its alignment metric reduces uncertainty to a scalar correction term via a variance-penalized inner product, rather than modeling the full distributional structure between modalities. These methods uniformly rely on deterministic or shallow probabilistic alignment, leaving the heterogeneous distributional gap between visual and textual representations insufficiently addressed.
 
\subsection{Uncertainty Estimation and Probabilistic Representations}
Uncertainty estimation in deep learning encompasses two complementary types: aleatoric uncertainty originating from inherent data noise that is irreducible regardless of training data volume, and epistemic uncertainty arising from limited model knowledge that can be progressively reduced with additional data \cite{kendall2017uncertainties}. Classical approaches such as MC Dropout \cite{gal2016dropout} and Deep Ensemble \cite{lakshminarayanan2017simple} estimate predictive uncertainty through post-hoc model-level inference via multiple stochastic forward passes or multiple independently trained models, but lack explicit modeling of uncertainty at the feature representation level. Recent work proposes density-based methods to disentangle both uncertainty types within a single forward pass \cite{mukhoti2023deep}, offering a more principled approach to feature-level uncertainty estimation. In cross-modal settings, images and text inherently exhibit one-to-many correspondences that deterministic point-to-point alignment cannot capture \cite{chun2021probabilistic}. PCME \cite{chun2021probabilistic} addresses this by modeling features as Gaussian distributions for probabilistic cross-modal matching, while ProbVLM \cite{upadhyay2023probvlm} estimates distributions over frozen VLM embeddings in a post-hoc manner. However, these probabilistic global matching methods remain insufficient for dense prediction tasks under distribution shift, and how to explicitly incorporate uncertainty into cross-modal feature alignment remains an open challenge.

\section{Methodology}

\subsection{Overview of DistMedVL}
We present the DistMedVL framework in Fig.~\ref{fig:architecture}(a). Given a medical image and its paired textual description, both inputs are processed by the frozen vision and text encoders of CLIPSeg~\cite{lueddecke22_cvpr}. To enable cross-modal probabilistic alignment at multiple semantic levels, we insert lightweight PCM-Adapters at layers $\{2, 4, 6, 8\}$ of both encoders. As illustrated in Fig.~\ref{fig:architecture}(b), each PCM-Adapter comprises a compression layer, two sequential modules that are the Mahalanobis Alignment Module (MAM) for text-to-vision probabilistic matching and the Distribution Flow Module (DFM) for vision-guided textual distribution refinement, and an expansion layer. The updated visual features are fed into a fine-tuned CLIPSeg decoder for segmentation prediction, while both modality features are additionally projected into a shared embedding space for contrastive alignment.
\subsection{Probabilistic Cross-Modal Adapter}
Existing vision-language adapters~\cite{dhakal2024vlsm} typically insert independent modules into each modality encoder without explicit cross-modal interaction, leaving the alignment to be implicitly learned from downstream supervision alone. To achieve explicit probabilistic alignment while preserving the pre-trained encoder representations, we devise the PCM-Adapter as a bottleneck-style module that operates in a compressed feature space. At each selected layer $\ell$, the intermediate visual features $\mathbf{F}_v^{(\ell)} \in \mathbb{R}^{N \times C}$ and textual features $\mathbf{F}_t^{(\ell)} \in \mathbb{R}^{L \times K}$ are first compressed and then restored via linear projections:
\begin{equation}
\begin{aligned}
    \mathbf{F}_v^{(\ell)\downarrow} &= \mathbf{F}_v^{(\ell)} \mathbf{W}_v^{\downarrow}, \quad
    \tilde{\mathbf{F}}_v^{(\ell)} = \hat{\mathbf{F}}_v^{(\ell)} \mathbf{W}_v^{\uparrow}, \\
    \mathbf{F}_t^{(\ell)\downarrow} &= \mathbf{F}_t^{(\ell)} \mathbf{W}_t^{\downarrow}, \quad
    \tilde{\mathbf{F}}_t^{(\ell)} = \hat{\mathbf{F}}_t^{(\ell)} \mathbf{W}_t^{\uparrow},
\end{aligned}
\label{eq:proj}
\end{equation}
 
\noindent where $\mathbf{W}_v^{\downarrow} \in \mathbb{R}^{C \times M}$, $\mathbf{W}_t^{\downarrow} \in \mathbb{R}^{K \times M}$ are compression projections that map both modalities into a shared $M$-dimensional space, and $\mathbf{W}_v^{\uparrow} \in \mathbb{R}^{M \times C}$, $\mathbf{W}_t^{\uparrow} \in \mathbb{R}^{M \times K}$ are expansion projections that restore the original dimensionality. This bottleneck design not only reduces computational overhead but also filters out redundant feature dimensions before probabilistic alignment. Within this compressed space, the MAM and DFM are applied sequentially:
\begin{equation}
\begin{aligned}
    \hat{\mathbf{F}}_v^{(\ell)} &= \text{MAM}^{(\ell)}\!\left(\mathbf{F}_v^{(\ell)\downarrow},\; \mathbf{F}_t^{(\ell)\downarrow}\right), \\
    \hat{\mathbf{F}}_t^{(\ell)} &= \text{DFM}^{(\ell)}\!\left(\hat{\mathbf{F}}_v^{(\ell)},\; \mathbf{F}_t^{(\ell)\downarrow}\right),
\end{aligned}
\label{eq:adapter}
\end{equation}
 
\noindent where the output of the MAM, $\hat{\mathbf{F}}_v^{(\ell)}$, is directly used as the visual input to the DFM at the same layer, as illustrated in Fig.~\ref{fig:architecture}(b). This sequential coupling is motivated by the observation that DFM benefits from receiving uncertainty-refined visual features rather than raw encoder outputs, enabling cascaded cross-modal refinement. The restored features $\tilde{\mathbf{F}}_v^{(\ell)}$ and $\tilde{\mathbf{F}}_t^{(\ell)}$ subsequently replace the original features as inputs to the next transformer layer. In this way, the PCM-Adapter realizes a bottleneck-style probabilistic alignment that enriches both modalities with cross-modal uncertainty-aware information while seamlessly preserving the frozen encoder representations.

\subsection{Mahalanobis Alignment Module}
 
Deterministic cross-modal matching methods, such as cosine similarity, treat all feature dimensions equally and produce fixed-point alignment, making them susceptible to noisy or ambiguous dimensions that are prevalent in clinical imaging data. Unlike cosine similarity, which only measures angular proximity in the feature space, the Mahalanobis distance incorporates the covariance structure of the underlying distributions, allowing the matching criterion to adaptively scale each feature dimension according to its estimated reliability. To this end, we devise MAM that models textual tokens as Gaussian distributions and computes patch-text compatibility via Mahalanobis distance, enabling variance-conditioned anisotropic alignment that inherently downweights unreliable feature dimensions, as illustrated in Fig.~\ref{fig:framework}(a).
 
Given the compressed features $\mathbf{F}_v^{(\ell)\downarrow}$ and $\mathbf{F}_t^{(\ell)\downarrow}$, we first project both into a compact $H$-dimensional space via linear layers with GELU activation to obtain $\mathbf{V} \in \mathbb{R}^{N \times H}$ and $\mathbf{T} \in \mathbb{R}^{L \times H}$. Within this space, we further apply separate semantic extractors followed by LayerNorm to obtain modality-specific representations $\mathbf{Z}_v \in \mathbb{R}^{N \times D}$ and $\mathbf{Z}_t \in \mathbb{R}^{L \times D}$. The LayerNorm is applied along the feature dimension independently for each token, ensuring that subsequent variance estimation reflects genuine uncertainty rather than heterogeneous feature scales. The MAM then models the textual semantic representations as Gaussian distributions by estimating per-token means and variances:
\begin{equation}
\begin{aligned}
    \boldsymbol{\mu}_t &= W_\mu \mathbf{Z}_t \in \mathbb{R}^{L \times D}, \\
    \boldsymbol{\sigma}_t^2 &= \text{softplus}(W_\sigma \mathbf{Z}_t) \in \mathbb{R}^{L \times D},
\end{aligned}
\label{eq:text_gaussian}
\end{equation}
 
\noindent where $W_\mu, W_\sigma \in \mathbb{R}^{D \times D}$ are learnable transformations applied to each token independently, and $\text{softplus}(\cdot)$ ensures strictly positive variances. Unlike textual tokens, which carry relatively sparse semantic cues, visual patch embeddings encode substantially denser spatial information. Directly estimating per-patch variance would yield noisy and unreliable estimates due to the redundancy and ambiguity inherent in dense visual features. We instead aggregate all patch embeddings into a global visual representation $\overline{\mathbf{Z}}_v = \frac{1}{N}\sum_{n=1}^{N} \mathbf{Z}_v^n \in \mathbb{R}^{D}$ and estimate an image-level variance $\boldsymbol{\sigma}_v^2 = \text{softplus}(W_{\sigma_v} \overline{\mathbf{Z}}_v) \in \mathbb{R}^{D}$, where $W_{\sigma_v} \in \mathbb{R}^{D \times D}$ maps the global representation into a per-dimension visual uncertainty estimate. The bilateral variance is then obtained by fusing both sources:
\begin{equation}
    \boldsymbol{\sigma}^2 = \boldsymbol{\sigma}_{t}^2 + \text{sigmoid}(\boldsymbol{\theta}_\lambda) \odot \boldsymbol{\sigma}_v^2 + \epsilon,
\label{eq:bilateral_var}
\end{equation}
 
\noindent where $\boldsymbol{\theta}_\lambda \in \mathbb{R}^D$ is a learnable parameter that controls how much image-level uncertainty is incorporated into each dimension, and $\text{sigmoid}(\cdot)$ constrains the visual variance contribution to $(0, 1)$. Since $\boldsymbol{\sigma}_v^2 \in \mathbb{R}^{D}$ is a single image-level estimate while $\boldsymbol{\sigma}_{t}^2 \in \mathbb{R}^{L \times D}$ is per-token, $\boldsymbol{\sigma}_v^2$ is broadcast across all $L$ rows, allowing each token to share a common image-level uncertainty prior while retaining its own token-specific textual variance. $\epsilon$ is a small constant for numerical stability. Since the bilateral variance provides per-dimension estimates, the covariance of each textual Gaussian is naturally diagonal, eliminating the need for full matrix inversion. The squared Mahalanobis distance between the $n$-th visual patch and the $l$-th textual Gaussian thus reduces to:
\begin{equation}
    d_M^2(\mathbf{Z}_v^n, \boldsymbol{\mu}_t^l) = \frac{1}{\sqrt{D}} \sum_{d=1}^{D} \frac{(Z_{v,d}^n - \mu_{t,d}^l)^2}{\sigma_{d}^{2,l}},
\label{eq:mahalanobis}
\end{equation}
 
\noindent where the $\frac{1}{\sqrt{D}}$ factor normalizes the distance to prevent magnitude scaling with dimensionality. Dimensions with larger variance contribute less to the overall distance, effectively suppressing unreliable feature dimensions in the alignment. By replacing point-wise matching with distribution-level alignment, the MAM inherently accounts for aleatoric uncertainty through per-dimension variance weighting and mitigates the adverse effect of epistemic uncertainty, since underrepresented features naturally exhibit higher variance and are thus automatically downweighted. Computing Eq.~\ref{eq:mahalanobis} for all patch-token pairs yields the distance matrix $\mathbf{D}_M \in \mathbb{R}^{N \times L}$, which is converted into a probabilistic compatibility matrix via temperature-scaled softmax:
\begin{equation}
    \mathbf{A} = \text{softmax}\left(-\mathbf{D}_M / \tau\right) \in \mathbb{R}^{N \times L},
\label{eq:compatibility}
\end{equation}
 
\noindent where $\tau$ is a learnable temperature parameter constrained to $[\tau_{\min}, \tau_{\max}]$ for training stability, and the negation ensures that smaller distances yield higher compatibility scores. The resulting $\mathbf{A}$ serves as a soft assignment matrix that captures the probabilistic affinity between visual patches and textual tokens. Based on this assignment, a dense semantic prior $\mathbf{S} = \mathbf{A}\mathbf{T} \in \mathbb{R}^{N \times H}$ is generated for each visual patch through weighted aggregation over textual features. The semantic prior is then adaptively integrated into the visual representation via a gating mechanism:
\begin{equation}
    \hat{\mathbf{F}}_v^{(\ell)} = W_o\!\left(\mathbf{V} + g(\mathbf{V}) \odot \mathbf{S}\right),
\label{eq:gated_fusion}
\end{equation}
 
\noindent where $g(\cdot)$ is a two-layer MLP with GELU activation and sigmoid output that produces a gate $g(\mathbf{V}) \in \mathbb{R}^{N \times H}$, $\odot$ is element-wise multiplication, and $W_o \in \mathbb{R}^{H \times M}$ projects the fused representation back to the compressed dimensionality. The residual connection ensures that the original visual representation $\mathbf{V}$ is preserved as the base, while the element-wise gating $g(\mathbf{V}) \odot \mathbf{S}$ allows each patch to selectively absorb relevant textual semantics. When the textual prior is unreliable or irrelevant, the gate outputs approach zero, effectively bypassing the semantic prior and preserving the original visual features. In this way, the MAM replaces deterministic point-to-point matching with variance-aware distribution-level alignment, where unreliable feature dimensions are naturally downweighted through the bilateral uncertainty encoded in the Mahalanobis distance.

\subsection{Distribution Flow Module}
 
Rather than applying MAM in reverse for the vision-to-text direction, we design a separate Distribution Flow Module to account for the granularity asymmetry between the two modalities, as illustrated in Fig.~\ref{fig:framework}(b). In the text-to-vision direction, visual patches can be sparsely matched to a small set of semantically relevant text tokens, making fine-grained probabilistic alignment tractable. However, reversing this formulation would require each text token to attend over a large number of spatially dense patches, resulting in diffuse attention maps that fail to capture discriminative visual cues. The DFM therefore adopts a global semantic aggregation strategy that distills both modalities into compact distributional summaries. However, aggregating dense visual patches into a single global representation raises a concern: the reliability of this summary depends on whether the patches reach semantic consensus. The DFM therefore incorporates a variance-driven reliability mechanism that assesses inter-patch semantic consistency and adaptively modulates the visual contribution accordingly.
 
Given the vision-enhanced features $\hat{\mathbf{F}}_v^{(\ell)}$ from the MAM and the original textual features $\mathbf{F}_t^{(\ell)\downarrow}$, we project both into a compact $K$-dimensional semantic space via linear projections $W_v, W_t \in \mathbb{R}^{M \times K}$ to obtain $\mathbf{D}_v$ and $\mathbf{D}_t$, which encourages more compact semantic aggregation. We then aggregate over visual patches to obtain a global semantic mean $\boldsymbol{\mu}_{v,g}$ and an inter-patch variance $\boldsymbol{\sigma}^2_{v,g}$ that quantifies the degree of semantic disagreement among patches along each feature dimension. A reliability gate transforms this variance into per-dimension reliability scores:
\begin{equation}
\begin{aligned}
    \boldsymbol{\mu}_{v,g} &= \textstyle\frac{1}{N}\sum_{i=1}^{N} \mathbf{D}_v^{(i)}, \quad
    \boldsymbol{\sigma}^2_{v,g} = \textstyle\frac{1}{N}\sum_{i=1}^{N} \left(\mathbf{D}_v^{(i)} - \boldsymbol{\mu}_{v,g}\right)^2, \\
    \boldsymbol{\gamma} &= \text{sigmoid}\!\left(\mathcal{G}\!\left(\boldsymbol{\sigma}^2_{v,g}\right)\right) \in \mathbb{R}^{K},
\end{aligned}
\label{eq:reliability}
\end{equation}
 
\noindent where $\mathbf{D}_v^{(i)} \in \mathbb{R}^{K}$ denotes the projected embedding of the $i$-th visual patch, and $\mathcal{G}(\cdot)$ is a two-layer MLP with GELU activation. The global mean $\boldsymbol{\mu}_{v,g}$ captures the overall semantic content of the image, while the global variance $\boldsymbol{\sigma}^2_{v,g}$ measures the inter-patch semantic consistency along each dimension. The gate is designed by default to high reliability at initialization, and learns during training to assign lower reliability scores to dimensions with higher inter-patch variance, adaptively suppressing uncertain visual signals before they propagate into the textual refinement. The semantic strength parameters for both modalities are then estimated as:
\begin{equation}
\begin{aligned}
    \boldsymbol{\beta}_v &= \left[\text{softplus}\!\left(W_{\beta_v} \boldsymbol{\mu}_{v,g}\right) + \epsilon \right] \odot \boldsymbol{\gamma}, \\
    \boldsymbol{\beta}_t &= \text{softplus}\!\left(W_{\beta_t} \boldsymbol{\mu}_{t,g}\right) + \epsilon,
\end{aligned}
\label{eq:strength}
\end{equation}
 
\noindent where $\boldsymbol{\mu}_{t,g} = \frac{1}{L}\sum_{j=1}^{L}\mathbf{D}_t^{(j)} \in \mathbb{R}^{K}$ denotes the global textual mean, $W_{\beta_v}, W_{\beta_t} \in \mathbb{R}^{K \times K}$ are learnable projections, and $\text{softplus}(\cdot)$ ensures strictly positive values. The element-wise multiplication with $\boldsymbol{\gamma}$ suppresses $\boldsymbol{\beta}_v$ in dimensions where visual patches are semantically inconsistent, preventing unreliable visual signals from corrupting the textual refinement. The refined textual semantic strength is obtained via a residual distribution flow:
\begin{equation}
    \boldsymbol{\beta}_t^* = \boldsymbol{\beta}_t + \text{MLP}_f\!\left(\boldsymbol{\beta}_v \| \boldsymbol{\beta}_t\right),
\label{eq:flow}
\end{equation}
 
\noindent where $\|$ denotes concatenation and $\text{MLP}_f(\cdot)$ is a two-layer MLP with GELU activation that learns a residual correction in the concentration parameter space. The residual formulation ensures that the visual distribution exerts a controlled adjustment rather than replacing the textual distribution entirely. The refined parameter is then projected back and broadcast across all $L$ text tokens to produce the final vision-guided textual feature:
\begin{equation}
    \hat{\mathbf{F}}_t^{(\ell)} = \mathbf{F}_t^{(\ell)\downarrow} + \text{MLP}_r\!\left(\boldsymbol{\beta}_t^*\right),
\label{eq:df_output}
\end{equation}
 
\noindent where $\text{MLP}_r(\cdot)$ projects $\boldsymbol{\beta}_t^*$ back into the original feature space and the offset is broadcast uniformly across all $L$ text tokens. This uniform broadcast reflects the global nature of the vision-to-text refinement that, rather than producing token-specific corrections, DFM applies a shared distributional shift, which adjusts all text tokens based on the aggregated visual evidence. In this way, DFM enables adaptive vision-guided refinement of textual distributions, where only reliable visual evidence with high inter-patch semantic consensus contributes to the cross-modal alignment, ensuring the robust distributional correction under varying input uncertainty.

\subsection{The DistMedVL Architecture Designs}
 
DistMedVL builds upon CLIPSeg~\cite{lueddecke22_cvpr}, a vision-language architecture pre-trained on natural image-text pairs. The framework adopts the ViT-B/16 vision encoder $\mathcal{E}_v$ and the Transformer-based text encoder $\mathcal{E}_t$, both consisting of 12 transformer layers, as frozen feature extractors. Given a medical image $\mathbf{X}_v$ and its paired textual description $\mathbf{X}_t$, the encoders produce visual features $\mathbf{F}_v \in \mathbb{R}^{N \times C}$ and textual features $\mathbf{F}_t \in \mathbb{R}^{L \times K}$, where $N$ and $L$ denote the number of visual patches and text tokens, and $C$ and $K$ the corresponding feature dimensionalities. The PCM-Adapters are inserted at layers $\{2, 4, 6, 8\}$ to perform cross-modal probabilistic alignment at multiple semantic levels, while only the adapter parameters and the CLIPSeg decoder are updated during training.
 
The final features $\mathbf{F}_v$ and $\mathbf{F}_t$ are fed into the CLIPSeg decoder $\mathcal{D}$, which is fine-tuned during training to generate segmentation predictions. To maintain global cross-modal alignment throughout training, we additionally project $\mathbf{F}_v$ and $\mathbf{F}_t$ into a shared embedding space via global average pooling followed by learnable linear projections for contrastive learning. The training objective combines a segmentation loss and a contrastive loss:
\begin{equation}
\begin{split}
    \mathcal{L}_\text{seg} &= \textstyle\frac{1}{2}\!\left(\mathcal{L}_\text{BCE} + \mathcal{L}_\text{Dice}\right),\\
    \mathcal{L}_{\rm DistMedVL} &= \lambda_1 \cdot \mathcal{L}_\text{seg} + \lambda_2 \cdot \mathcal{L}_\text{con},
\end{split}
\label{eq:total_loss}
\end{equation}
 where $\mathcal{L}_\text{con}$ follows the CLIP contrastive formulation~\cite{radford2021learning}, and $\lambda_1 = 0.5$, $\lambda_2 = 0.1$ are loss weighting coefficients following MedCLIPSeg~\cite{koleilat2026medclipseg}. The entire framework introduces only 6.3M trainable parameters from the PCM-Adapters and decoder, while the frozen encoders contribute the remaining parameters. To rigorously evaluate the generalization capability of the proposed framework, no data augmentation is employed during training beyond image resizing, ensuring that performance gains are attributable to the model architecture rather than augmentation strategies.

\section{Experiments}

\subsection{Datasets}
We evaluate DistMedVL across 8 benchmark datasets spanning four imaging modalities. Five datasets, which are TN3K~\cite{gong2022thyroid}, BUSI~\cite{ALDHABYANI2020104863}, ISIC-2016~\cite{gutman2016skin}, Kvasir-SEG~\cite{jha2019kvasir} and Covid19~\cite{degerli2022osegnet},are used for training and evaluation. Three additional datasets, which are BUID~\cite{ardakani2023open}, CVC-ClinicDB~\cite{tajbakhsh2015automated} and PKTN~\cite{sun2024cliptnsegmultimodalhybridframework}, serve exclusively as out-of-distribution targets for domain generalization. The dataset splits are summarized in Table~\ref{tab:dataset}. For BUSI, ISIC-2016, Kvasir-SEG, Covid19, BUID and CVC-ClinicDB, we adopt the dataset splits and text prompt annotations from MedCLIPSeg~\cite{koleilat2026medclipseg}. For TN3K and PKTN, text descriptions are generated using Qwen2.5~\cite{hui2024qwen2}, guided by dataset descriptions and ground-truth masks.

\begin{table*}[!t]
\caption{Dataset splits for data efficiency and domain generalization experiments. The top five datasets are used for training and evaluation under varying data ratios. The bottom three datasets serve exclusively as out-of-distribution targets for domain generalization, where all available samples are used for testing.}
\label{tab:dataset}
\centering
\begin{tabular}{lcccccc}
\hline
Dataset & Train 10\% & Train 25\% & Train 50\% & Train 100\% & Validation & Test \\
\hline
TN3K \cite{gong2022thyroid}      & 259 & 647  & 1295 & 2591 & 288  & 614  \\
BUSI \cite{ALDHABYANI2020104863}       & 62  & 156  & 312  & 624  & 78   & 78   \\
ISIC-2016 \cite{gutman2016skin}  & 80  & 202  & 405  & 810  & 90   & 379  \\
Kvasir-SEG \cite{jha2019kvasir} & 80  & 200  & 400  & 800  & 100  & 100  \\
Covid19 \cite{degerli2022osegnet}    & 571 & 1429 & 2858 & 5716 & 1429 & 2113 \\
BUID \cite{ardakani2023open}       & -   & -    & -    & -    & -    & 232  \\
CVC-ClinicDB \cite{tajbakhsh2015automated}  & -   & -    & -    & -    & -    & 612  \\
PKTN \cite{sun2024cliptnsegmultimodalhybridframework}       & -   & -    & -    & -    & -    & 1005 \\
\hline
\end{tabular}
\end{table*}

\begin{table*}[!t]
\caption{Data efficiency and epistemic uncertainty analysis. Models are trained with 10\%, 25\%, 50\%, and 100\% of the training data to evaluate segmentation performance under varying levels of epistemic uncertainty induced by limited supervision. Results are averaged across 5 datasets.}
\label{tab:comparison}
\centering
\begin{tabular}{lcccccccc}
\hline
\multirow{2}{*}{Model} &
  \multicolumn{2}{c}{10\%} &
  \multicolumn{2}{c}{25\%} &
  \multicolumn{2}{c}{50\%} &
  \multicolumn{2}{c}{100\%} \\ 
  \cmidrule(lr){2-3}
  \cmidrule(lr){4-5}
  \cmidrule(lr){6-7}
  \cmidrule(lr){8-9}
 &
  Dice &
  NSD &
  Dice &
  NSD &
  Dice &
  NSD &
  Dice &
  NSD \\ \hline
 \rowcolor[gray]{0.92}
 \multicolumn{9}{c}{Unimodal Approaches}   \\ 
 UNet \cite{ronneberger2015u} & 62.40 & 66.81 & 70.26 & 74.61 & 75.64 & 80.04 & 80.38 & 84.65 \\
 DCSAU-Net \cite{xu2023dcsau} & 60.06 & 64.72 & 67.46 & 71.79 & 75.87 & 80.30 & 80.05 & 84.61  \\
 FAT-Net \cite{wu2022fat} & 73.95 & 78.42 & 80.09 & 84.41 & 82.19 & 86.39 & 84.51 & 88.70 \\
 Swin-Unet \cite{cao2022swin} & 75.39 & 80.70 & 80.09 & 84.67 & 82.70 & 87.07 & 84.25 & 88.63   \\
 Trans-Unet \cite{chen2021transunet} & 77.11 & 81.93 & 78.78 & 83.47 & 82.24 & 86.82 & 83.42 &  87.90  \\
 CFFormer \cite{li2026cfformer} & 76.10 & 81.12 & 79.35 & 83.89 & 82.40 & 86.65 & 84.51 & 88.67 \\ 
 \rowcolor[gray]{0.92}
 \multicolumn{9}{c}{Crossmodal Approaches} \\ 
 TGANet \cite{tomar2022tganet} & 76.68 & 81.24 & 81.75 & 86.19 & 84.36 & 88.60 & 85.93 & 90.39 \\ 
 LViT \cite{li2023lvit} & 68.16 & 72.63 & 77.72 & 82.22 & 81.24 & 85.68 & 84.09 & 88.39 \\
 CLIPSeg \cite{luddecke2022image} & 77.05 & 82.49 & 80.66 & 85.74 & 83.29 & 88.46 & 84.76 & 89.80\\
 LanGuideMedSeg \cite{zhong2023ariadne} & 77.31 & 81.79 & 79.28 & 83.47 & 84.77 & 89.44 & 86.45 & 91.17   \\
 VLSM-Adapter \cite{dhakal2024vlsm} & 79.65 & 84.79 & 82.25 & 87.38 & 83.42 & 86.72 & 84.91 & 90.20 \\
 RecLMIS \cite{huang2024cross} & 71.64 & 76.27 & 77.96 & 82.36 & 82.42 & 86.85 & 85.69 & 90.01 \\
 CausalCLIPSeg \cite{chen2024causalclipseg} & 72.03 & 76.72 & 79.17 & 83.83 & 83.59 & 88.07 & 85.90 & 90.37\\
 MedCLIPSeg \cite{koleilat2026medclipseg} & 78.27 & 83.38 & 83.06 & 87.99 & 85.01 & 89.78 & 86.44 & 91.07 \\
 DistMedVL (Ours) & \textbf{81.50} & \textbf{86.54} & \textbf{84.63} & \textbf{89.44} & \textbf{86.22} & \textbf{90.91} & \textbf{87.63} & \textbf{92.36}\\
\hline
\end{tabular}
\end{table*}

\subsection{Implementation Details}
All experiments are conducted on a single NVIDIA RTX A6000 GPU (48GB). We use a batch size of 24 and optimize the model using the AdamW optimizer with a learning rate of $3 \times 10^{-4}$ and default betas $(0.9, 0.999)$. The learning rate is scheduled by cosine annealing with a minimum of $1 \times 10^{-4}$ over 100 epochs. Input images are resized to $224 \times 224$ and normalized using ImageNet statistics. No data augmentation is applied during training or testing.
 
\subsection{Evaluation Metrics}
We adopt two complementary metrics to comprehensively assess segmentation quality. The Dice Similarity Coefficient (DSC) measures the volumetric overlap between the predicted segmentation mask $P$ and the ground truth $G$, and it is particularly sensitive to region-level accuracy. The Normalized Surface Distance (NSD) evaluates boundary quality by computing the fraction of predicted boundary points that lie within a specified tolerance distance $\tau$ of the ground-truth boundary, providing a complementary measure of contour precision that is clinically relevant for delineating anatomical structures with complex or ambiguous boundaries. We report both metrics as percentages, where higher values indicate better segmentation accuracy and boundary delineation, respectively.

\begin{figure*}[!t]
\centering
\includegraphics[width=0.9\textwidth]{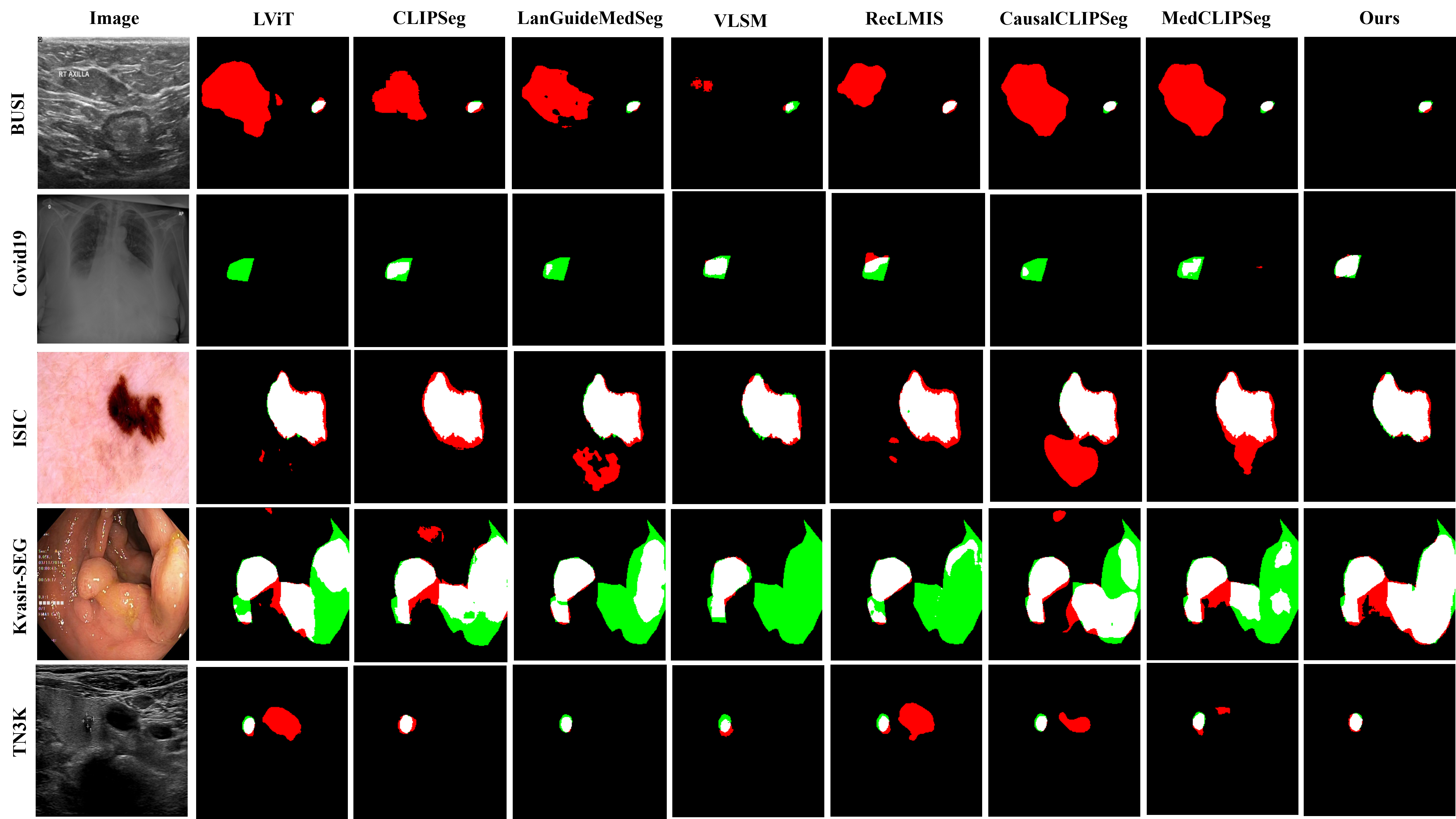}
\caption{Qualitative comparison of vision-language models on medical image segmentation under the 100\% training data setting. White regions indicate correctly segmented areas, green regions denote under-segmentation, and red regions denote over-segmentation.}
\label{fig:visual_all}
\end{figure*}
\begin{table*}[h]
\caption{Domain generalization results. Models are trained on the source domain and evaluated on out-of-distribution target datasets. Three groups of experiments are conducted under varying levels of aleatoric uncertainty.}
\label{tab:domain_generalization}
\centering
\resizebox{\textwidth}{!}{\begin{tabular}{l cc cc cc cc cc cc cc}
\toprule
\multirow{3}{*}{Method}
  & \multicolumn{4}{c}{Breast Ultrasound}
  & \multicolumn{4}{c}{Colonoscopy}
  & \multicolumn{4}{c}{Thyroid Nodule} \\
\cmidrule(lr){2-5}\cmidrule(lr){6-9}\cmidrule(lr){10-13}
  & \multicolumn{2}{c}{Source: BUSI}
  & \multicolumn{2}{c}{Target: BUID}
  & \multicolumn{2}{c}{Source: Kvasir-SEG}
  & \multicolumn{2}{c}{Target: CVC-Clinic}
  & \multicolumn{2}{c}{Source: TN3K}
  & \multicolumn{2}{c}{Target: PKTN} \\
\cmidrule(lr){2-3}\cmidrule(lr){4-5}\cmidrule(lr){6-7}\cmidrule(lr){8-9}\cmidrule(lr){10-11}\cmidrule(lr){12-13}
  & Dice & NSD & Dice & NSD & Dice & NSD & Dice & NSD & Dice & NSD & Dice & NSD \\ \hline
\rowcolor[gray]{0.92}
\multicolumn{13}{c}{Unimodal Approaches} \\
UNet \cite{ronneberger2015u} & 70.67 & 74.86 & 71.44 & 76.01 & 82.13 & 85.34 & 57.96 & 61.95 & 81.08 & 86.49 & 46.06 & 51.67 \\
DCSAU-Net \cite{xu2023dcsau} & 77.28 & 81.65 & 69.37 & 74.18 & 74.37 & 78.09 & 50.09 & 54.33 & 80.70 & 86.12 & 47.76 & 53.61 \\
FAT-Net \cite{wu2022fat} & 80.37 & 84.44 & 77.41 & 82.35 & 89.23 & 92.34 & 79.08 & 83.28 & 81.94 & 87.18 & 56.90 & 63.53 \\ 
Swin-Unet \cite{cao2022swin} & 78.64 & 83.07 & 73.11 & 77.44 & 89.55 & 92.61 & 79.04  & 83.76 & 81.97 & 87.57 & 60.28 & 68.34 \\ 
TransUnet \cite{chen2021transunet} & 75.91 & 80.18 & 77.42 & 82.35 & 89.80 & 93.62 & 81.43 & 86.52 & 80.92 & 86.61 & 54.86 & 62.38 \\
CFFormer \cite{li2026cfformer} & 79.46 & 83.61 & 75.19 & 79.65 & 89.43 & 92.44 & 79.70 & 84.04 & 81.47 & 86.63 & 60.32 & 67.86 \\
\rowcolor[gray]{0.92}
\multicolumn{13}{c}{Crossmodal Approaches} \\
TGANet \cite{tomar2022tganet} & 82.26 & 86.37 & 79.77 & 84.50 & 90.00 & 93.57 & 80.15 & 84.31 & 82.47 & 88.43 & 61.65 & 69.27 \\
LViT \cite{li2023lvit} & 81.42 & 85.86 & 64.37 & 69.60 & 87.08 & 90.43 & 68.16 & 73.76 & 80.71 & 85.91 & 54.10 & 60.27 \\
CLIPSeg \cite{luddecke2022image} & 81.88 & 86.82 & 55.81 & 61.23 & 86.92 & 90.46 & 81.09 & 86.67 & 80.56 & 86.72 & 63.02 & 72.84 \\ 
LanGuideMedSeg \cite{zhong2023ariadne} & 84.65 & 88.72 & 81.77 & 86.76 & 90.19 & 93.76 & 83.60 & 87.94 & 82.53 & 88.64 & 65.64 & 73.28 \\ 
VLSM-Adapter \cite{dhakal2024vlsm} & 80.96 & 86.55 & 79.53 & 84.78 & 89.61 & 93.76 & 82.50 & 88.09 & 80.65 & 86.92 & 56.78 & 66.08 \\ 
RecLMIS \cite{huang2024cross} & 83.71 & 88.09 & 68.78 & 73.22 & 87.36 & 90.46 & 76.10 & 80.63 & 82.09 & 87.25 & 57.67 & 64.54 \\ 
CausalCLIPSeg \cite{chen2024causalclipseg} & 83.43 & 87.92 & 76.82 & 81.72 & 90.41 & 93.79 & 79.17 & 83.93 & 81.67 & 87.13 & 57.52 & 64.96 \\
MedCLIPSeg \cite{koleilat2026medclipseg} & 84.32 & 88.89 & 82.66 & 87.27 & 90.05 & 93.72 & 82.35 & 87.63 & 82.37 & 88.34 & 63.49 & 72.15\\ 
DistMedVL (Ours)    & \textbf{86.01} & \textbf{90.65} & \textbf{83.43} & \textbf{88.20} & \textbf{90.98} & \textbf{94.78} & \textbf{84.64} & \textbf{89.67} & \textbf{84.39} & \textbf{90.26} & \textbf{68.02} & \textbf{77.48} \\ \hline
\end{tabular}}
\end{table*}

\begin{figure*}[!h]
  \centering
  \includegraphics[width=0.80\linewidth]{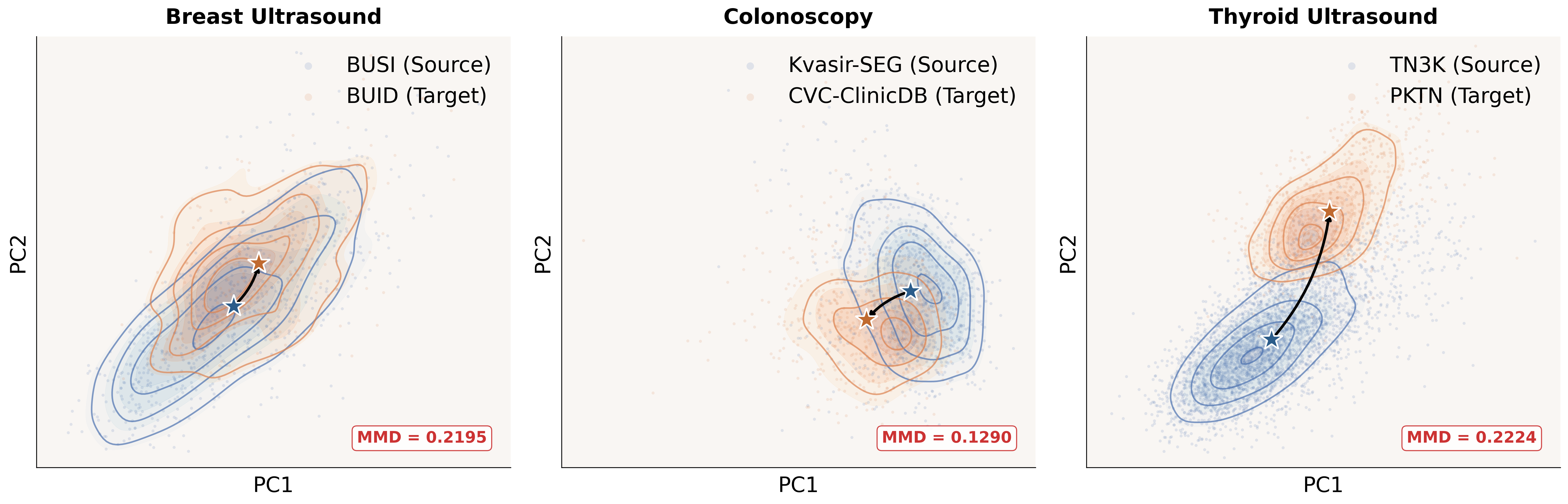}
  \caption{Visualization of domain shift between source and target datasets via PCA projection of extracted features. The Maximum Mean Discrepancy (MMD) quantifies the distributional discrepancy between domains. Arrows indicate the direction of distribution shift from source to target centroids.}
  \label{fig:domain}
\end{figure*}

\subsection{Comparison with State-of-the-Art Methods}
 
We compare DistMedVL with both unimodal approaches (UNet~\cite{ronneberger2015u}, DCSAU-Net~\cite{xu2023dcsau}, FAT-Net~\cite{wu2022fat}, Swin-Unet~\cite{cao2022swin}, TransUNet~\cite{chen2021transunet}, CFFormer~\cite{li2026cfformer}) and crossmodal approaches (TGANet~\cite{tomar2022tganet}, LViT~\cite{li2023lvit}, CLIPSeg~\cite{luddecke2022image}, LanGuideMedSeg~\cite{zhong2023ariadne}, VLSM-Adapter~\cite{dhakal2024vlsm}, RecLMIS~\cite{huang2024cross}, CausalCLIPSeg~\cite{chen2024causalclipseg}, MedCLIPSeg~\cite{koleilat2026medclipseg}). To simultaneously assess data efficiency and epistemic uncertainty, we train all models with 10\%, 25\%, 50\% and 100\% of the training data, since reducing the training set size naturally increases epistemic uncertainty while keeping aleatoric uncertainty constant within the same distribution~\cite{kendall2017uncertainties}. Evaluations are averaged across the five benchmark datasets.
 
As shown in Table~\ref{tab:comparison}, DistMedVL consistently outperforms all competing methods across all data regimes. At full supervision (100\% training data), our method achieves 87.63\% Dice and 92.36\% NSD, surpassing the best-competing crossmodal approach MedCLIPSeg~\cite{koleilat2026medclipseg} by 1.19\% in Dice and 1.29\% in NSD, and outperforming the strongest unimodal method by over 3\% in both metrics. Under high epistemic uncertainty with only 10\% training data, the performance gap widens further, with DistMedVL achieving at least 1.85\% Dice and 1.75\% NSD improvement over the best-competing approach. This widening advantage under data scarcity suggests that probabilistic alignment becomes increasingly beneficial as epistemic uncertainty grows, since the variance-conditioned matching inherently downweights unreliable features that arise from insufficient training. The consistent gains across all data ratios demonstrate that our approach provides robust benefits regardless of the epistemic uncertainty level.

Fig.~\ref{fig:visual_all} presents qualitative segmentation results across the five benchmark datasets. When competing methods produce fragmented predictions or fail to accurately localize pathological regions, DistMedVL retains precise lesion capture and boundary delineation. This is attributed to the variance-aware alignment in the PCM-Adapter that suppresses unreliable feature dimensions during cross-modal matching, reducing the impact of both imaging noise and model uncertainty on the final prediction.

\subsection{Domain Generalization}
 
To further evaluate robustness against epistemic uncertainty arising from distribution shift, we conduct cross-domain experiments where models are trained on a source domain and directly evaluated on unseen target domains without fine-tuning. We select three source-target pairs representing progressively increasing domain shift: breast ultrasound (BUSI~\cite{ALDHABYANI2020104863} $\rightarrow$ BUID~\cite{ardakani2023open}), colonoscopy (Kvasir-SEG~\cite{jha2019kvasir} $\rightarrow$ CVC-ClinicDB~\cite{tajbakhsh2015automated}), and thyroid ultrasound (TN3K~\cite{gong2022thyroid} $\rightarrow$ PKTN~\cite{sun2024cliptnsegmultimodalhybridframework}). As illustrated in Fig.~\ref{fig:domain}, PCA projections of CLIP-extracted features confirm the ordering that the breast ultrasound pair exhibits substantial distributional overlap, the colonoscopy pair shows moderate divergence, while the thyroid pair displays the most significant separation with the highest MMD value, indicating the most severe epistemic uncertainty.
 
As shown in Table~\ref{tab:domain_generalization}, DistMedVL consistently outperforms all competing methods across all three domain shift settings. In the breast ultrasound transfer (BUSI $\rightarrow$ BUID), our model surpasses MedCLIPSeg~\cite{koleilat2026medclipseg} on the target domain by 0.77\% in Dice and 0.93\% in NSD. In the colonoscopy transfer (Kvasir-SEG $\rightarrow$ CVC-ClinicDB), DistMedVL achieves 84.64\% Dice on the target domain, maintaining a 1.04\% Dice lead over the best-competing method. In the most challenging thyroid transfer (TN3K $\rightarrow$ PKTN), where epistemic uncertainty is highest as evidenced by the largest MMD in Fig.~\ref{fig:domain}, our model achieves a substantial 2.38\% Dice and 4.20\% NSD advantage over the state-of-the-art. Notably, the performance advantage of DistMedVL grows as the domain shift intensifies, demonstrating that probabilistic alignment becomes increasingly beneficial under higher epistemic uncertainty. This property is particularly relevant for real-world clinical deployment, where models inevitably encounter unseen imaging protocols and patient populations. Rather than degrading gracefully like deterministic baselines, DistMedVL appears to leverage the additional uncertainty signal to recalibrate its predictions, suggesting that explicit distributional modeling offers a more robust foundation for generalization than point-estimate representations alone.

\begin{figure}[t]
  \centering
  \includegraphics[width=\linewidth]{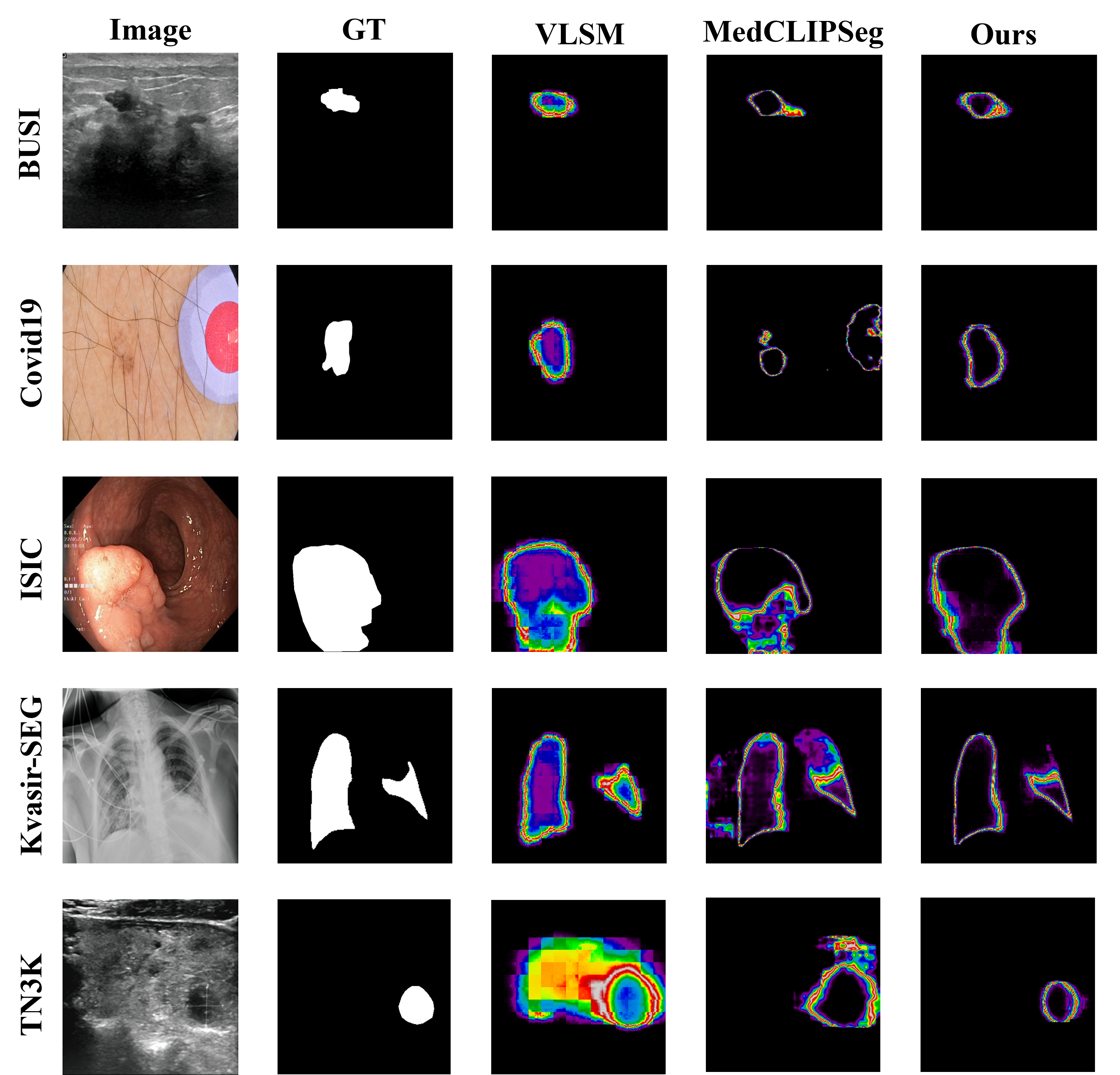}
  \caption{Qualitative comparison of uncertainty maps. Warmer colors indicate higher predictive uncertainty, while cooler colors denote lower uncertainty.}
  \label{fig:uncertainty}
\end{figure}

\subsection{Perturbation Robustness}
 
To evaluate robustness against aleatoric uncertainty, we introduce controlled perturbations to both input modalities: Gaussian blur ($7 \times 7$ kernel, $\sigma = 2$) for images, and spatial orientation inversion for text prompts. We evaluate three scenarios: image-only, text-only, and joint perturbation. We compare against MedCLIPSeg~\cite{koleilat2026medclipseg}, which also explicitly addresses uncertainty, on TN3K~\cite{gong2022thyroid} (high intrinsic speckle noise) and Kvasir-SEG~\cite{jha2019kvasir} (relatively clean data) to examine degradation across varying levels of intrinsic aleatoric uncertainty.
 
As shown in Table~\ref{tab:pur}, DistMedVL exhibits substantially smaller performance degradation under all perturbation scenarios. Under image perturbation on TN3K, our Dice decreases by only 1.08\%, compared to 2.44\% for MedCLIPSeg. On Kvasir-SEG, our degradation under image perturbation is also lower than that of MedCLIPSeg. The advantage is more pronounced under text perturbation, where our model benefits from the MAM's distribution-level alignment that downweights high-variance dimensions regardless of the query state, unlike MedCLIPSeg's query-dependent variance penalty that weakens when the query itself is corrupted. The DFM further reinforces robustness through its reliability gate, which detects perturbation-induced inter-patch inconsistency and suppresses unreliable visual signals. Under joint perturbation where aleatoric uncertainty is maximized, DistMedVL continues to maintain significantly lower degradation across both datasets, demonstrating the resilience of our dual uncertainty-aware alignment against aleatoric uncertainty across both modalities.
\begin{table}[t]
\centering
\caption{Perturbation experiment results. IP: Gaussian blur applied to input images. TP: spatial directional cues in text prompts inverted. \ding{51}/\ding{55} indicate presence/absence of perturbation.}
\label{tab:pur}
\renewcommand{\arraystretch}{1.15}
\setlength{\tabcolsep}{5pt}
\begin{tabular}{lcccccc}
\toprule
\multirow{2}{*}{Model} & \multirow{2}{*}{IP} & \multirow{2}{*}{TP} & \multicolumn{2}{c}{TN3K} & \multicolumn{2}{c}{Kvasir-SEG} \\
\cmidrule(lr){4-5} \cmidrule(lr){6-7}
 &  &  & Dice & NSD & Dice & NSD \\
\hline
MedCLIPSeg \cite{koleilat2026medclipseg} & \ding{55} & \ding{55} & 82.37 & 88.34 & 90.05 & 93.72 \\
MedCLIPSeg \cite{koleilat2026medclipseg}  & \ding{51} & \ding{55} & 79.93 & 85.63 & 89.54 & 93.23 \\
MedCLIPSeg \cite{koleilat2026medclipseg}  & \ding{55} & \ding{51} & 76.96 & 82.26 & 89.31 & 93.06 \\
MedCLIPSeg \cite{koleilat2026medclipseg}  & \ding{51} & \ding{51} & 74.82 & 80.10 & 89.15 & 92.91 \\
\hline
DistMedVL (Ours) & \ding{55} & \ding{55} & 84.39 & 90.26 & 90.98 & 94.78\\
DistMedVL (Ours) & \ding{51} & \ding{55} & 83.31 & 88.97 & 90.58 & 94.45 \\
DistMedVL (Ours)  & \ding{55} & \ding{51} & 80.83 & 86.42 & 90.74 & 94.53\\
DistMedVL (Ours) & \ding{51} & \ding{51} & 79.00 & 84.49 & 90.23 & 94.03 \\
\bottomrule
\end{tabular}
\end{table}

\subsection{Ablation Study}
 
Table~\ref{tab:ablation} presents the ablation study of the PCM-Adapter components. The decoder-only baseline without adapter insertion achieves 81.12\% Dice and 86.28\% NSD, as the frozen encoder pretrained on natural images lacks domain-specific priors for medical image segmentation. Inserting the MAM alone improves Dice to 85.76\%, while the DFM alone yields 85.02\%, confirming the effectiveness of both modules in bridging the domain gap through intermediate feature adaptation. Combining both modules in a parallel configuration further improves performance on BUSI (86.96\% Dice), demonstrating their complementary contributions. However, the sequential configuration, where the DFM receives uncertainty-refined features from the MAM, achieves the best overall performance across both datasets, particularly on Kvasir-SEG (90.98\% Dice vs.\ 90.35\% in parallel). This validates our design motivation: the MAM first produces uncertainty-refined visual features, providing a more reliable input for the subsequent DFM to perform global semantic alignment, whereas the parallel configuration operates on raw features independently and cannot benefit from this cascaded refinement.
\begin{table}[t]
\centering
\caption{Ablation study on adapter configurations within the PCM Adapter. \ding{51}/\ding{55} denote the presence/absence of each module. Seq. indicates sequential arrangement.}
\label{tab:ablation}
\renewcommand{\arraystretch}{1.15}
\setlength{\tabcolsep}{5pt}
\begin{tabular}{ccccccc}
\toprule
\multirow{2}{*}{MDM} & \multirow{2}{*}{DFM} & \multirow{2}{*}{Sequential} & \multicolumn{2}{c}{BUSI} & \multicolumn{2}{c}{Kvasir-SEG} \\
\cmidrule(lr){4-5} \cmidrule(lr){6-7}
 &  &  & Dice & NSD & Dice & NSD \\
\hline

\ding{55} & \ding{55} & \ding{55} & 81.12 & 86.28 & 87.94 & 91.77 \\
\ding{51} & \ding{55} & \ding{55} & 85.76 & 90.52 & 90.21 & 93.98 \\
\ding{55} & \ding{51} & \ding{55} & 85.02 & 90.55 & 89.68 & 93.35 \\
\ding{51} & \ding{51} & \ding{55} & 86.96 & 90.67 & 90.35 & 94.11 \\
\ding{51} & \ding{51} & \ding{51} & 86.01 & 90.65 & 90.98 & 94.78 \\
\bottomrule
\end{tabular}
\end{table}
Fig.~\ref{fig:uncertainty} presents predictive uncertainty maps estimated via Monte Carlo Dropout with 30 forward passes. Compared to VLSM-Adapter~\cite{dhakal2024vlsm} and MedCLIPSeg~\cite{koleilat2026medclipseg}, DistMedVL consistently concentrates high-entropy regions along segmentation boundaries while suppressing spurious uncertainty in both foreground and background. This indicates well-calibrated posterior estimates and demonstrates that the PCM-Adapter preserves fine-grained boundary delineation where competing methods produce spatially diffuse uncertainty.

\section{Conclusion}
In this paper, we propose DistMedVL, a probabilistic vision‑language framework for uncertainty‑aware medical image segmentation. We introduce a lightweight PCM‑Adapter consisting of two sequential modules, Mahalanobis Alignment and Distribution Flow, which explicitly mitigate the impact of uncertainty on segmentation through variance‑conditioned cross‑modal alignment and reliability‑gated distributional refinement. To thoroughly evaluate its robustness against uncertainty, we conduct experiments on data efficiency, cross‑domain generalization, and perturbation robustness. The results demonstrate that DistMedVL effectively suppresses uncertainty across diverse scenarios. 
\bibliographystyle{IEEEtran}
\bibliography{ref}

\vfill

\end{document}